%% file: main.tex
\documentclass[sigconf]{acmart}
\def\BibTeX{{\rm B\kern-.05em{\sc i\kern-.025em b}\kern-.08emT\kern-.1667em\lower.7ex\hbox{E}\kern-.125emX}}

\usepackage[font=small]{caption}
\usepackage{verbatim}
\usepackage{amsmath}
\usepackage{amssymb}
\usepackage{booktabs} 
\usepackage{algorithm}
\usepackage{algcompatible}
\usepackage{algpseudocode}
\usepackage{multirow}
\usepackage{color}
\usepackage{threeparttable}
\usepackage{pifont}
\usepackage{caption}
\usepackage{subcaption}
\usepackage[unicode]{hyperref}

\usepackage{tikz}
\usepackage{calc}

\renewcommand\footnotetextcopyrightpermission[1]{} 

\begin{document}

%

\title{
\vspace{-26pt} NAIS: Neural Architecture and Implementation Search and its Applications in Autonomous Driving
}

\affiliation{ \normalsize
Cong Hao$^{1,4}$, Yao Chen$^{3}$, Xinheng Liu$^{1}$, Atif Sarwari$^{2}$, Daryl Sew$^{2}$, Ashutosh Dhar$^{1,4}$,
Bryan Wu$^{2}$, Dongdong Fu$^{2}$, Jinjun Xiong$^{4,1}$, Wen-mei Hwu$^{1,4}$, Junli Gu$^{2}$ and Deming Chen$^{1,4}$}
\affiliation{ \normalsize
	\institution{$^1$University of Illinois at Urbana-Champaign, $^2$XMotors.ai}
%
	\institution{$^3$Advanced Digital Sciences Center, Singapore, $^4$IBM-Illinois Center for Cognitive Computing Systems Research (C$^3$SR)}
}

\begin{abstract}
The rapidly growing demands for powerful AI algorithms 
in many application domains have motivated massive investment in both high-quality deep neural network (DNN) models and high-efficiency implementations.
In this position paper, we argue that a simultaneous DNN/implementation co-design methodology, named \textbf{N}eural \textbf{A}rchitecture and \textbf{I}mplementation \textbf{S}earch \textbf{(NAIS)},
deserves more research attention to boost the development productivity and efficiency of both DNN models and implementation optimization. We propose a stylized design methodology that can drastically cut down the search cost while preserving the quality of the end solution.
As an illustration, we discuss this DNN/implementation methodology in the context of both FPGAs and GPUs.
We take autonomous driving as a key use case as it is one of the most demanding areas for high quality AI algorithms and accelerators. We discuss how such a co-design methodology can impact the autonomous driving industry significantly. We identify several research opportunities in this exciting domain.
\end{abstract}

\maketitle
\input{sections/01-introduction.tex}
\input{sections/02-codemethod.tex}

\input{sections/03-dodesign-fpga.tex}

\input{sections/04-codeautodrive.tex}
\input{sections/05-experiments.tex}
\input{sections/06-related.tex}

\input{sections/07-conandfuwork.tex}

\bibliographystyle{unsrt}
\bibliography{ref}

\end{document}

%% file: sections/01-introduction.tex
\section{introduction}

\begin{figure}
    \centering
    \includegraphics[width=0.45\textwidth]{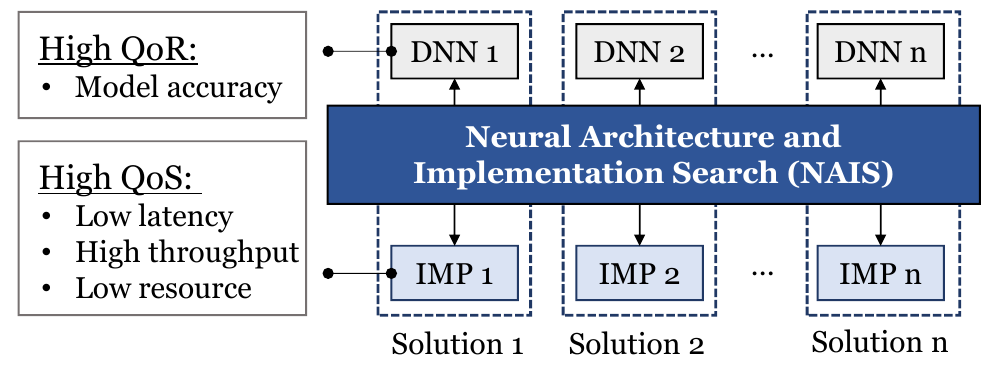}
    \caption{ \textbf{N}eural \textbf{A}rchitecture and \textbf{I}mplementation \textbf{S}earch~\textbf{(NAIS)}
   generates DNNs and optimizes their implementations simultaneously, achieving both high Quality of Result (QoR) and Quality of Service (QoS).}
    \label{fig:co-design}
\end{figure}

\begin{figure*}
    \centering
    \includegraphics[width=0.93\textwidth]{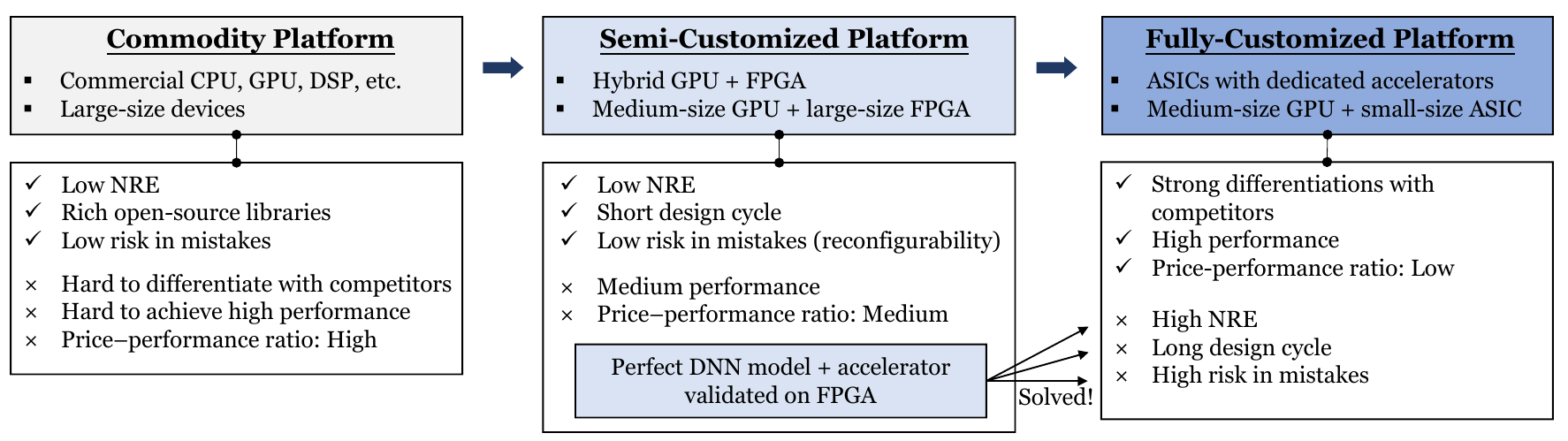}
    \caption{Industrial position of reconfigurable architectures (e.g. FPGA) in autonomous driving: an important step from semi-customized platform to fully-customized platform.}
    \label{fig:fpga-in-industry}
    \vspace{-6pt}
\end{figure*}

The world has seen tremendous improvements in AI algorithms as well as their high performance implementations in recent years.
Remarkable achievements have been demonstrated for AI algorithms in many areas 
with expeditious improvements in algorithm quality and robustness.
Deep neural network (DNN) is one of the most popular AI algorithms with impressive advancements, from AlexNet \cite{krizhevsky2012imagenet} to modern models \cite{szegedy2015going,simonyan2014very,he2016deep}.
Meanwhile, the optimization techniques for high performance implementations of AI algorithms on hardware are also being intensively studied.
Such implementation techniques include kernel and DNN optimizations on GPUs and TPUs \cite{guerreiro2015multi, TensorRT, cuDNN, edgetpu}, 
accelerator designs on customizable hardware such as FPGAs~\cite{zhang2015optimizing,sharma2016high,zhang2018dnnbuilder,zhang2018caffeine,clouddnn} and AI chips \cite{chen2016eyeriss,yin2017high}.

Despite many of these accomplishments, there are still many challenges, one of which is the gap between high quality DNN models during design and their implementation performance during deployment.
One reason for such a gap is isolated design of DNNs and optimization of their implementations, where the former does not integrate sufficient hardware knowledge, and the later does not have enough freedom to accommodate pre-designed DNNs at such a late stage. 
Instead, DNNs and their hardware implementations need to be designed simultaneously, i.e., \textit{DNN/implementation co-design}, as illustrated in Fig.~\ref{fig:co-design}. We call it \textbf{N}eural \textbf{A}rchitecture and \textbf{I}mplementation \textbf{S}earch \textbf{(NAIS)}.
The outputs of NAIS include both DNNs that are of high quality of result (QoR), and implementations that are of high quality of service (QoS).
The NAIS methodology brings immense optimization opportunities for:
\begin{itemize}
    \item {
    Proposing specific hardware-oriented DNN models. For DNN deployment, there are many hardware candidates such as GPUs, cloud and edge TPUs, cloud and embedded FPGAs, each of which has largely different characteristics such as computation capability, memory capacity and bandwidth.
    The NAIS method will explore DNNs based on specific hardware features and search for DNNs with the best match.
    }
    \item {
    Meeting resource and performance constraints. The NAIS method will search for DNNs within available hardware resources and performance constraints, which provides predictable and guaranteed performance for DNN deployment.
    }
    \item {
    Shortening design cycles. While existing top-down design methods require back-and-forth efforts to find satisfying solutions, an automated NAIS flow can simultaneously find an optimized DNN model and its deployment on hardware.
    }

\end{itemize}


In modern industry applications, as AI algorithms are increasingly adopted, high performance computing platforms are in great need, especially with reconfigurable devices for acceleration.
Take autonomous driving as an example, which is one of the most demanding areas for high QoR AI algorithms and high performance computing implementation.
Fig.~\ref{fig:fpga-in-industry} shows three types of computing platforms for autonomous driving: commodity platform composed of commercial CPUs, GPUs or DSPs, semi-customized platform composed of GPUs and FPGAs, and fully-customized platform composed of dedicated ASICs.
As shown in the figure, though fully-customized platforms are most favorable in terms of high performance and low price-performance ratio (e.g. \$/Gops), they suffer from high non-recurring engineering (NRE) cost, long design cycle and high risk in making mistakes, which hinders their wide adoption.
In contrast, with reconfigurable devices such as FPGAs, semi-customized platforms become a competitive alternative with a good trade-off in performance and cost.
Moreover, once the AI algorithms and their hardware implementations
have been fully validated on FPGAs, the design can be made into ASICs to take advantage of what a fully-customized platform can offer.
Thus, finding high quality AI algorithms with their optimized implementations on reconfigurable devices not only provides good solutions for semi-customized platforms, but also provides a good path to move from semi to fully customized platforms.
Because of this, there is a pressing need for NAIS, an automatic co-design of AI algorithms and their optimized implementations, on GPUs, FPGAs and ASICs, given the widely varying device characteristics and the large design space of both algorithmic and implementation optimization.

Motivated by those opportunities,
in this work, we propose NAIS as a simultaneous DNN/implementation co-design approach to
effectively search for high quality DNN models and high performance implementations for different hardware platforms.
We demonstrate how such a NAIS approach can be utilized to solve real-world applications, including
autonomous driving.

%% file: sections/02-codemethod.tex
\section{NAIS Design Methdology}

\begin{figure}[t]
    \centering
    \begin{subfigure}[]{0.43\textwidth}
        \includegraphics[width=\textwidth]{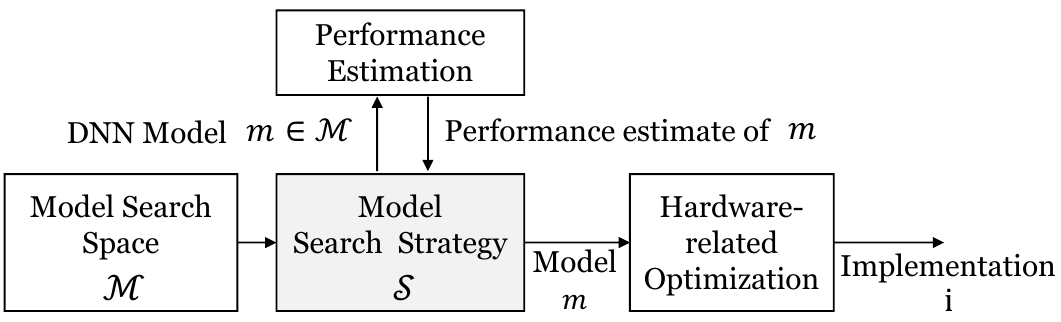}
        \caption{}
        \label{fig:old-nas}
    \end{subfigure}
    \begin{subfigure}[]{0.43\textwidth}
        \includegraphics[width=\textwidth]{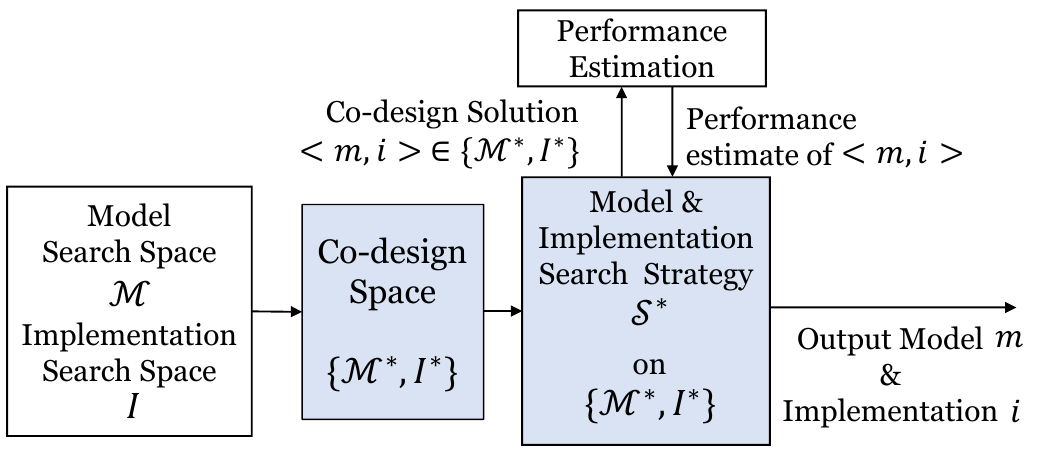}
        \caption{}
        \label{fig:new-nas}
    \end{subfigure}
    \caption{(a) The traditional neural architecture search (NAS) \cite{elsken2019neural}. (b) Our proposed NAIS co-design methodology.}
    \vspace{-12pt}
\end{figure}

\begin{figure*}
    \centering
    \includegraphics[width=0.95\textwidth]{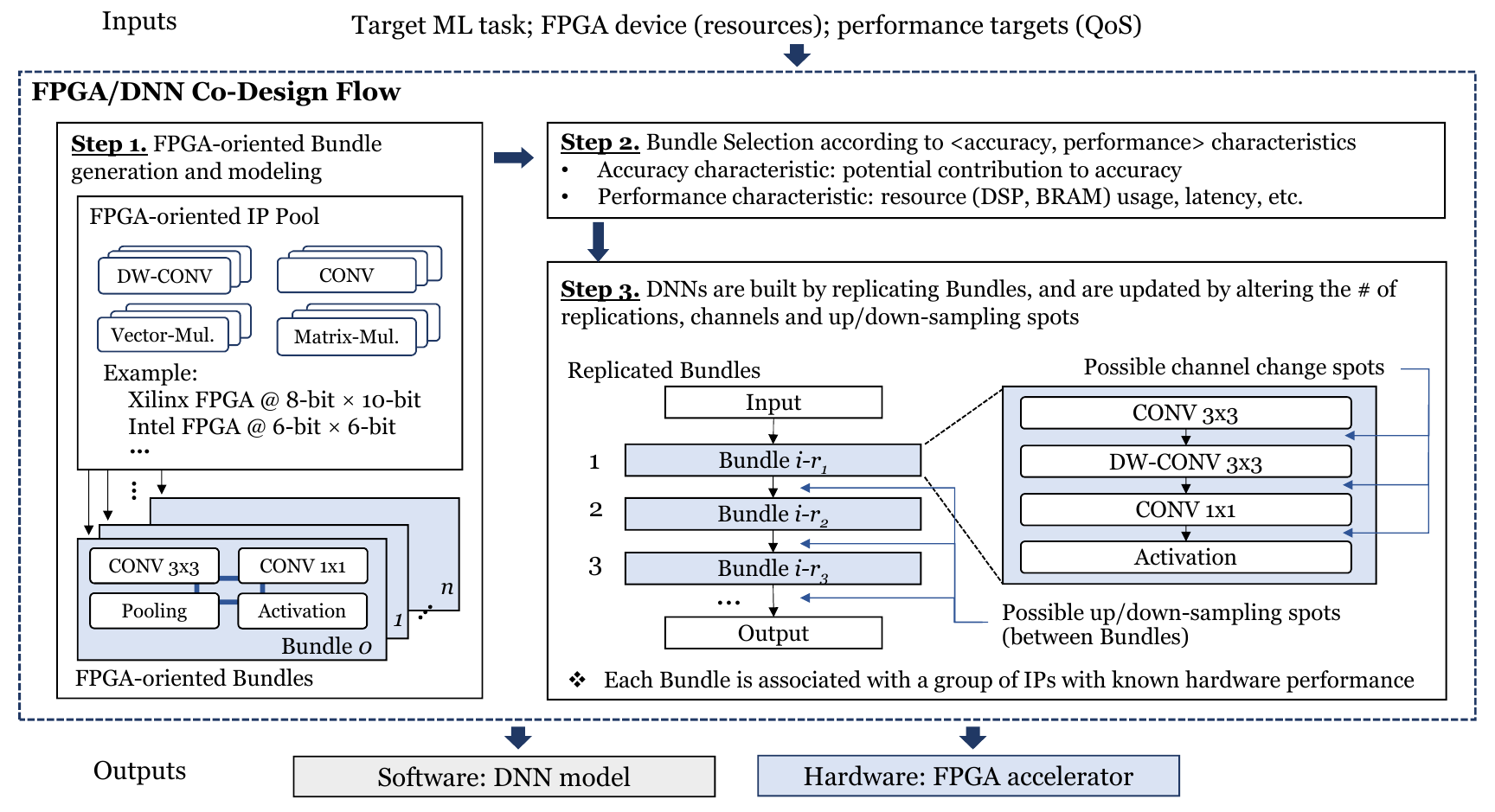}
    \caption{NAIS applied to DNN/FPGA co-design.}
    \label{fig:overall}
\end{figure*}

A NAIS methodology has two tasks: to search for DNNs of high QoR (e.g. accuracy), and for implementations of high QoS (e.g. latency, throughput).
Such an implementation can be an optimized software stack on a given accelerator device such as GPUs, or a customized hardware accelerator on FPGAs, CGRAs, and ASICs.

\textbf{Neural Architecture Search (NAS)}.
For DNN search, most existing NAS engines can find high quality DNNs.
As illustrated in Fig.~\ref{fig:old-nas},
given a model search space $\mathcal{M}$, a NAS engine applies a certain search strategy $\mathcal{S}$ such as reinforcement learning or evolutionary algorithm. During the search, the performance (QoR) of the model $m\in \mathcal{M}$ is estimated and provided back to the NAS engine.
After NAS generates a satisfying DNN,
it will be implemented and deployed on GPU, FPGA or other devices.
During the search, however, implementation optimization is not considered.
For example in a recent hardware-aware NAS approach \cite{cai2018proxylessnas}, it considers directly measured inference latency on the GPU but does not explore optimization techniques. This will result in a large performance gap between estimation and final implementation, especially when there are multiple candidate devices, each requiring different optimization techniques. 
When targeting FPGAs, it becomes more important that DNN search and implementation search being tightly coupled during NAS: different accelerator implementation configurations can result in large performance variation.

\textbf{Neural Architecture and Implementation Search (NAIS)  --- Beyond NAS}.
To fully explore implementation optimizations and to consider the impacts of implementation on DNNs, we propose a fully simultaneous DNN/implementation co-design approach:
it not only searches for neural architectures, but also searches for implementation optimizations, i.e., a \textbf{N}eural \textbf{A}rchitecture and \textbf{I}mplementation \textbf{S}earch, \textbf{NAIS}.
As illustrated in Fig.~\ref{fig:new-nas}, the NAIS search space includes both model search space $\mathcal{M}$ and implementation search space $\mathcal{I}$.
We combine the two spaces as a co-design space $\{\mathcal{M^*, I^*}\}$, and apply a joint search strategy $\mathcal{S^*}$ on the co-design space.
During NAIS, each solution $(m, i)$ is composed of two parts: a DNN model solution $m$, and a corresponding implementation solution $i$, where specific optimization techniques have been applied to $i$.
After searching, the NAIS engine outputs both the DNN model and its optimized hardware implementation.
The design space of NAIS is the product of the design space of DNN search and the design space of implementation optimization, which can be huge. 
Such combined design space makes the co-design procedure time-consuming and hard to converge. Innovative research is needed to address this new challenge.

In this position paper, we first prototype a NAIS methodology in the context of DNN/FPGA co-design, and show how we effectively narrow the co-design space to generate high quality DNNs and their FPGA implementations within the resource constraints of a target FPGA.
We then discuss how such a NAIS design methodology can be extended for GPU in a similar fashion.

%% file: sections/03-dodesign-fpga.tex

\section{nais for fpga}

\subsection{DNN/Implementation Co-design Space}
\label{sec:co-design-space}

The FPGA accelerator optimization problem is very complicated and requires comprehensive domain-specific knowledge. For example, the overall accelerator architecture (pipelined or folded), the number of IPs and parallelism of each IP, data quantization, buffer allocation, data reuse, etc., and each has a significant impact on the final performance.
Besides, the FPGA underlying characteristics (DSP structure, block RAM, bandwidth, etc.) and available resources can be very different between FPGA devices or families.

To efficiently narrow down the combined design space of NAIS for a target FPGA, we propose to co-design both DNN structure and its FPGA accelerator implementation using hardware-aware basic building blocks, named \textbf{Bundles} \cite{hao2019fpga}.
A Bundle represents a set of sequential DNN layers, 
and a DNN can be constructed by replicating a Bundle for $n$ times with configurations (the '\textbf{A}' in N\textbf{A}IS).
Meanwhile, a Bundle is composed of a set of FPGA configurable IPs, where each IP is well designed and highly optimized, and the Bundle is used to construct the FPGA implementation (the '\textbf{I}' in NA\textbf{I}S).
For DNN, each Bundle replication can be configured to have different number of channels of its layers;
for FPGA, a Bundle can be configured to have a certain number of IP instances, and each IP instance with specific parallel factors, data precision, on-chip buffers, etc.
When a Bundle is selected and configured,
both the DNN model and its accelerator can be determined. 
That is, Bundles provide a stylized approach to design both the DNNs and FPGA implementations, thus narrowing the search space efficiently.

\subsection{Overall Co-Design Flow}
\label{sec:overall-co-design-flow}

Given the co-design space,
Fig.~\ref{fig:overall} shows our proposed NAIS co-design flow targeting FPGA \cite{hao2019fpga}.
The inputs include a machine learning task such as image classification or object detection, resource constraints of a specific FPGA device, and performance target such as frame rate.
The outputs include both DNN models and corresponding FPGA accelerator with achieved performance.
Inside the co-design flow, there are three major steps.

\textbf{Step 1: FPGA-oriented Bundle generation}.
First, we design a pool of FPGA-oriented IPs considering specific FPGA characteristics such as DSP and BRAM structures.
The IPs may have same functionality but different designs.
For example, to best utilize the DSP resource, a Xilinx FPGA may best support 8-bit $\times$ 10-bit multiplication IPs, while an Intel FPGA may best support 9-bit $\times$ 9-bit multiplication IPs.
Based on the IPs, we build FPGA-oriented Bundles, where the data tiling, pipelining and data movement between these IPs are considered.

    
\textbf{Step 2: Bundle selection}.
Second, we apply Bundle evaluation to reduce the co-design space by only selecting the most promising Bundles for future exploration.
Each Bundle will be evaluated regarding its resource utilization and potential contribution to DNN accuracy.
We build a Bundle-wise DNN template with fixed front-end and back-end structures, and insert one Bundle (with replications) in the middle each time \cite{hao2019fpga, zhang2019skynet}. Such Bundle-wise DNNs will be quickly trained using a small number of epochs to evaluate the accuracy.
The Bundles on the resource-accuracy Pareto curve will be selected.
 
\textbf{Step 3: Hardware-aware DNN search and update}. 
Third, we perform hardware-aware DNN search.
The inputs include the initial DNNs, performance objectives such as latency, and resource constraints.
We use stochastic coordinate descent (SCD) to update three variables related to DNN structure: the number of Bundle replications; down-sampling configuration between Bundles; and the number of channels in each Bundle. During the iterations of SCD, only DNNs within the resource constraints and performance requirements are kept for downstream training.
In such a way, the final generated DNNs are more structured, resulting in more efficient hardware implementations.

\subsection{FPGA-oriented IP design}

Since the FPGA's characteristics vary with device vendors and types, a well designed IP must fully consider such characteristics to achieve the maximum performance while minimizing resource utilization.
We discuss two most important factors as an illustration: the structure of DSPs and embedded block memory.

\subsubsection{DSP consideration}

Table~\ref{tab:dsp-device} shows different multiplication and accumulation precision of DSPs in different FPGA devices, where the variation can be large even within the same vendor.
The computational IPs should be carefully designed based on the underlying DSP structure to take full advantage of its computation capability, which, in turn, affects the DNN design.

One important factor that must be considered is DNN's data precision.
Take the Xilinx DSP48E1 and DSP48E2 as examples.
Assume a simple case of two multiplications, $a\times c$ and $b\times c$ with a common multiplier $c$, and $a$, $b$, $c$ have $b_a$, $b_b$, $b_c$ bits, respectively. To increase multiplication parallelism, one possibility is to let two multiplicands (in this case $a$ and $b$) occupy one DSP input $I_1$, and let the common multiplier (in this case $c$) occupy the other input $I_2$, so that the two multiplications can be conducted at one clock cycle. To ensure correctness, there must be at least $b_c$ empty bits between $a$ and $b$, so that 
the two products do not overlap with each other in the output.
When using DSP48E1, which supports 18-bit $\times$ 25-bit multiplications and 48-bit accumulation, if $a$ and $b$ are both 8-bit and occupy the 25-bit operand, then $c$ must not exceed 9-bit ($8+8+9 \leq 25$);
when using DSP48E2, $c$ can be 10-bit ($8+8+10 \leq 27$).
In a scenario where $a$ and $b$ are activations and $c$ is the weight, if the target FPGA has DSP48E1, the DNN weights should be quantized to 9-bit or less, while with DSP48E2, the weights can be 10-bit.
Similarly, if the target device is Intel FPGA, the preferable quantization changes accordingly. For example, on Stratix V, <9-bit, 9-bit> is more preferred than <10-bit, 9-bit> for weights and activation, because Stratix V DSPs support $9 \times 9$-bit multiplications.

\begin{figure}[t]
    \centering
    \includegraphics[width=0.4\textwidth]{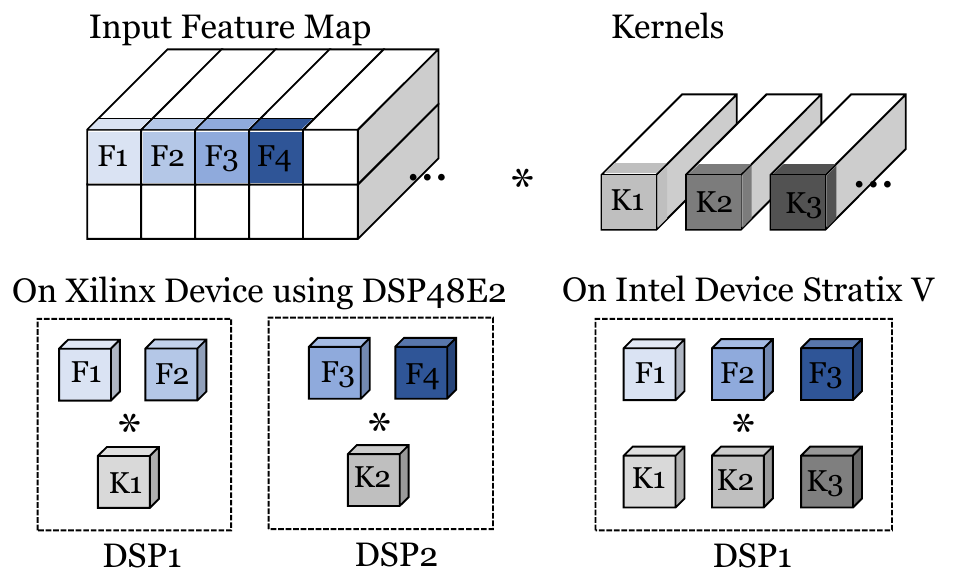}
    \caption{Different DSP structures lead to different IP designs.}
    \label{fig:dsp-ip-difference}
\end{figure}

Moreover, the DSP structure also affects the computation pattern and parallelism, which determines the detailed IP design.
Fig.~\ref{fig:dsp-ip-difference} shows an example of a convolution $1\times 1$ IP targeting Xilinx and Intel Arria V devices, respectively.
On Xilinx DSPs, to conduct two multiplications in parallel by sharing a common multiplier, one kernel will once consume two pieces of feature map data to fully utilize one DSP.
On Intel Arria V series, where one DSP is capable of running three independent $9 \times 9$ multiplications,
three kernels will consume three pieces of different feature map data at a time.
Such differences between DSPs will result in disparate IP designs and performance, and need to be considered in the NAIS engine.

\subsubsection{BRAM consideration}

\begin{table}[t]
\centering
\small
\def\arraystretch{0.9}
\caption{BRAM data width of different FPGA devices.}
\begin{tabular}{| c | c | c |}
\hline
     \multicolumn{2}{|c|}{Device} & Data Width \\ \hline
     Xilinx \cite{xilinxram} & RAMB18E1 & 1, 2, 4, 9, 18\\
            & RAMB36E1 & 1, 2, 4, 9, 18, 36\\ \hline
     Intel \cite{intelram} & MLAB & 8, 9, 10, 16, 18, 20 \\
            & M9K & 1, 2, 4, 8, 9, 16, 18, 32$^*$, 36$^*$ \\
            & M20K & 8, 10, 16, 20, 32, 40 \\
            & M144K & 8, 9, 16, 18, 32, 36, 64$^*$, 72$^*$ \\
            & eSRAM & 72 \\ \hline
    \multicolumn{3}{l}{\footnotesize *Only applicable for single-port RAM, simple-dual port RAM, and single-port ROM} \\
\end{tabular}
\label{tab:bram-width}
\vspace{-4pt}
\end{table}

%
%


On-chip block memory (BRAM) is another important design factor to consider. Effectively utilizing on-chip memory for data buffering can greatly reduce the amount of off-chip data movement, thus reducing both latency and energy consumption.
Table~\ref{tab:bram-width} shows the supported data width of different FPGA devices.
For Xilinx, the commonly used bit widths are 9, 18 and 36 in its RAMB18E1 and RAMB36E1.
For Intel, the common block memory is M20K, which has a capacity of 20Kb organized into either 10- or 20-bit storage words and read/write operations.
The on-chip data buffers need to be carefully allocated to align with the block memory depth and width.
For example, if a continuous buffer is allocated to be 21Kb, it will occupy two blocks of Intel M20k, resulting in a large waste of the second block.

The differences in block memory structure can affect the desirable DNN designs as well.
Take the buffer allocation for feature maps using Xilinx RAMB18E1 as an example.
If the input feature map dimension of a layer is $96 \times 96 \times 1$ (one channel) using 8-bit data, the number of occupied RAM blocks is 4, and a slightly larger feature map will consume an additional block.
Usually, the intermediate feature map dimensions are closely related to the original input size and up/down sampling.
Therefore, resizing the input image to $384\times 384~(96 \times 4)$ may be better than $388\times 388~(97 \times 4)$ as far as on-chip buffer allocation is concerned.

The discussions in this section show that the structures of DSPs and BRAMs play an important role in guiding the DNN design in a NAIS framework.

\begin{table}[t]
\centering
\small
\def\arraystretch{0.9}
\caption{Multiplication precision of different FPGA devices.}
\begin{tabular}{| c | c | c |}
\hline
     Device & Precision Mode & Accumulator \\ \hline
     Xilinx 7 series (DSP48E1) \cite{xilinxdsp48e1} & One 25 $\times$ 18 & 48-bit \\\hline
     Xilinx UltraScale (DSP48E2) \cite{xilinxdsp48e2} & One 27 $\times$ 18 & 48-bit \\\hline


                      
    Intel Stratix V \cite{intelStratixV} & Three 9 $\times$ 9 & 64-bit\\ 
                      & Two 18 $\times$ 18 & \\
                      & One 18 $\times$ 36 & \\
                      & One 27 $\times$ 27 & \\ \hline
                      
    Intel Arria V \cite{intelArriaV} & Three 9 $\times$ 9 & 64-bit\\
                    & Two 18$\times$18 & \\
                  & One 27$\times$27 & \\\hline
                  
    Intel Stratix 10 \cite{intelstratix10} & Two 18 $\times$ 19 & 64-bit\\
    Intel Arria 10 \cite{intelarria10} & One 27$\times$27 & \\\hline


\hline
\end{tabular}
\label{tab:dsp-device}
\vspace{-12pt}
\end{table}

\section{nais for gpu}
\begin{figure}
    \centering
    \includegraphics[width=0.5\textwidth]{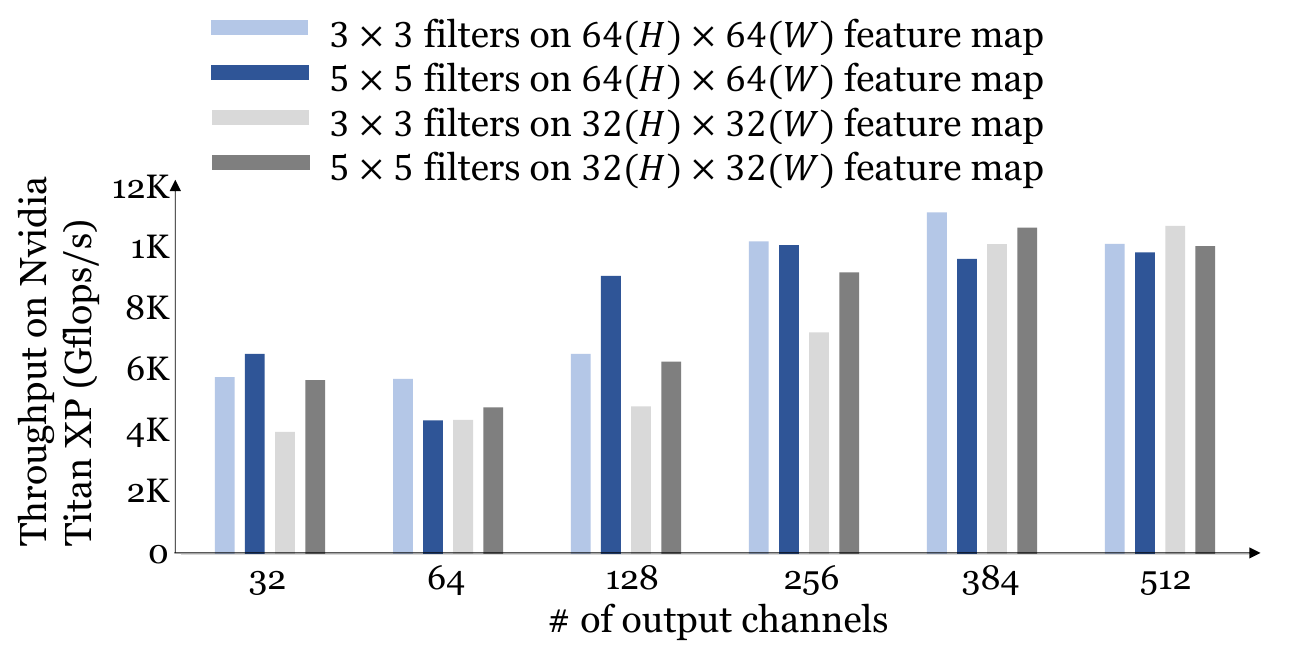}
    \caption{Different kernel size and feature map size result in different throughput on Nvidia Titan XP using cuBLAS 10.0.}
    \label{fig:cublas-diff}
    \vspace{-6pt}
\end{figure}

There are recent works discussing hardware-aware NAS targeting GPUs \cite{cheng2018searching, marculescu2018hardware}.
However, during NAS, GPU kernel configuration and optimization were ignored, which is a non-trivial problem that has attracted a lot of research interest \cite{zhou2017performance, guerreiro2015multi, tsai2016performance}.
Table \ref{tab:gpu_parameters} summarizes a set of GPU architecture-specific and kernel-specific parameters, which can affect the kernel configuration and performance on a specific GPU~\cite{guerreiro2015multi}.
These parameters vary greatly with different GPU generations.
Hence, the selection of the most adequate configurations of the GPU kernels has proven to be a difficult design optimization problem~\cite{guerreiro2015multi}.
In \cite{tsai2016performance}, it is demonstrated that for just a single AlexNet layer with 4 tunable parameters, the possible configurations are $17\times 254$, and the performance ranges from 44.7 to 5735.8 Gflop/s on an AMD Fury X GPU. 
Even with GPU optimization tools, such as TensorRT \cite{TensorRT} on top of cuDNN and cuBLAS, one kernel can still have varied performance.
Fig.\ref{fig:cublas-diff} shows the variation of GPU throughput when computing one convolution layer with different filter configurations.

To apply NAIS in DNN and GPU implementation co-design,
we can generalize the aforementioned DNN/FPGA co-design methodology.
For example,
a GPU-oriented Bundle can be defined as well.
One GPU Bundle is composed of a set of GPU kernels, which shall be configured and optimized targeting the specific GPU device
(usually the GPUs used for training and for inference are different).
The parameters of the Bundle may include favorable matrix shapes for a matrix-multiplication kernel, the number of threads, the batch size, etc.
Such Bundle optimization problem is being intensively studied with auto-tuning tools such as \cite{tsai2016performance} and \cite{guerreiro2015multi}, where \cite{guerreiro2015multi} especially targets multi-kernel optimizations.
With optimized kernel Bundles, structured NAS \cite{zoph2017learning} can be applied. Similar to the normal cells and reduction cells used in \cite{zoph2017learning}, we can search for DNNs with different configurations of normal Bundles and reduction Bundles, which are optimized GPU kernels in NAIS.
Leveraging both GPU Bundle optimization and structured NAS search, we can develop a NAIS engine that can be naturally applied to GPU and DNN co-design.

With more advanced profiling capabilities, such as a recent MLModelScope \cite{mlmodelscope} tool, we can easily evaluate and profile DNN models across different datasets, frameworks and hardware at scale and across stack. With such detailed layer-wise and kernel-wise profiling data,  roofline models for all kernels can be built to understand whether a kernel configuration is computation or memory bound. All those performance models can be leveraged to furtherance the development of NAIS for GPU.


\begin{table}[]
    \centering
    \small
    \def\arraystretch{0.9}
\caption{GPU architecture and kernel specific parameters~\cite{guerreiro2015multi}.}
    \begin{tabular}{|c|l|}
    \hline
        Type &  Parameter \\\hline
    Architecture &  Max. number of blocks per SM \\
    specific      &  Max. number of warps per SM \\
                  &  Shared memory per SM \\
                  &  Shared memory alloc. unit size \\
                  &  Max. number of registers per SM \\
                  &  Registers alloc. unit size \\
                  &  Warp size \\
                  &  Max. number of threads per SM \\\hline
                  
    Kernel        &  Number of warps per thread block \\
    specific     &  Shared memory per block \\
                  &  Number of registers per thread \\\hline
                  
    Architecture  & Max. number of thread blocks \\
    \& kernel specific  & Hardware utilization measure \\ \hline
    
    \end{tabular}
    \label{tab:gpu_parameters}
    \vspace{-12pt}
\end{table}

%% file: sections/04-codeautodrive.tex
\section{nais for autonomous driving}

 An autonomous driving system collects a large amount of data from surrounding environment, and executes a complicated software pipeline for localization, perception, prediction, planning and control.
 To support a safe and robust software pipeline, a powerful computing platform as well as high quality AI algorithms are indispensable,
 and a NAIS approach is imperative to support both.

\begin{figure*}
    \centering
    \includegraphics[width=\textwidth]{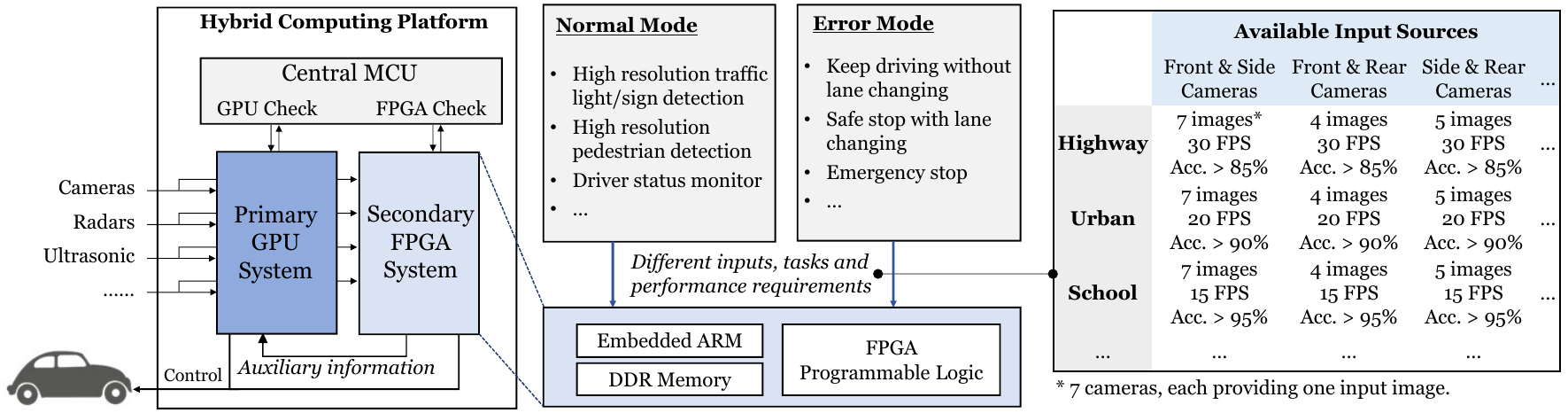}
    \caption{A hybrid GPU + FPGA system in autonomous driving \cite{hao2019hybrid}. The functionality, input sources and performance requirements for the FPGA system are complicated and largely vary with driving scenarios. }
    \label{fig:hybrid-system}
    \vspace{-6pt}
\end{figure*}

\vspace{-4pt}
\subsection{Computing Platforms}

Currently, GPUs are the prevailing computing platforms for autonomous driving with high programmablity, flexibility and performance.
In this demand, Nvidia brought Drive AGX \cite{nvidiadriveagx}, a powerful autonomous driving hybrid platform built on Nvidia Xavier, incorporating 8-core CPUs, deep learning accelerators (DLA), integrated GPU and 
programmable vision accelerators (PVA). Within these components, DLA is most adequate for DNN-based inference, which can be replaced by ASICs or FPGAs.
Therefore, for competitive differentiation, some leading autonomous driving companies have started to adopt specialized platforms.
For example, the Mobileye \cite{mobileye} and Tesla \cite{tesla} have developed their own chips to achieve outstanding AI performance and low power.
FPGAs, on the other hand, have also been a popular computing platform for autonomous driving cars because of its appealing advantages such as industrial reliability, specialization, high performance and low power.
There are ongoing efforts from technology companies and academia institutions for FPGA based solutions \cite{Nithin2014advanced, Okuda2014survey}.
Xilinx, for example, has developed their ADAS using Zynq-7000 SoC-based FPGA devices~\cite{xilinxadas}.
In a recent collaborative work of UIUC and XMotors \cite{hao2019hybrid},
a hybrid GPU + FPGA computing system for autonomous driving has been proposed.
Fig.~\ref{fig:hybrid-system} illustrates the hybrid system, where the GPU serves as a primary system, and the FPGA serves as a secondary system for failure fallback and providing auxiliary information for assistive driving.

Given the emerging needs of semi-customized platforms with reconfigurable devices and a full-customized platform as a direct next step, the DNN and implementation co-design is highly expected to boost the ongoing productivity and platform evolving.

\vspace{-8pt}
\subsection{Autonomous Driving Algorithms}

Self-driving is a comprehensive robotic capability including parking, driving and in-cabin intelligence functions, and each contains a set of varied sub-functions with different AI algorithms, input sources and performance requirements.

\textit{Varied sub-functions.}
Self-driving pipeline requires different functions and algorithms in different scenarios.
For example, the algorithms for parking and driving can be very different: parking task focuses on parking lot detection with near distance, localization and low speed vehicle control, while driving task focuses on motion objects, obstacle, lane detection within hundreds of meters and high speed vehicle control.
In-cabin intelligence also has multiple sub-functions such as DSM (Driver State Monitoring), voice recognition, gesture based interactions, and passenger detection.
Another example can be seen in the hybrid GPU and FPGA system proposed in \cite{hao2019hybrid}, shown in Fig. \ref{fig:hybrid-system}.
In the system error mode, the FPGA executes different tasks:
when the car is on a highway, it keeps driving and while maintaining a minimum speed limit;
when in urban area, it slows down the car and applies a safe pull-over.
Each scenario requires different DNNs to be mapped to FPGA.

\textit{Varied input sources.}
The self-driving system will be provided with varied input sources.
For example, parking functions usually use surrounding cameras and ultra-sonic sensors,
while highway drive uses multiple 
front, side and rear cameras with assistance of radars.
Another example is shown in Fig.~\ref{fig:hybrid-system}, where the FPGA accepts input images with different resolutions:
in normal mode, it may conduct traffic light detection using high resolution input images,
while in error mode, it runs simplified autonomous driving pipeline using low resolution input image for object and lane detection.

\textit{Varied performance requirements.}
Autonomous driving algorithms need to cope with numerous and complicated driving scenarios with different performance requirements.
For example, when driving in highways, the perception module requires at least 30 FPS but the number of objects to be detected may be limited to cars, lanes and traffic signs;
in urban area, it requires 20 FPS but with a larger number of objects to detect;
in school area, it may require 15 FPS but need a higher accuracy especially for pedestrian detection.


Given such variations in sub-functions, input sources and performance requirements, 
the detailed AI algorithms to each situation will be very different. 
Accordingly, the overall pipeline including other traditional algorithms will be significantly different, and all have to run on the same centralized electronic control unit (ECU) platform. 
Thus, an automatic NAIS co-design flow will enable us to explore the optimal solution under each situation.

%% file: sections/05-experiments.tex
\section{experiment results}

\begin{table}
\centering
\small
\def\arraystretch{0.9}
\caption{ Peek performance and accuracy under different data precisions of SkyNet \cite{zhang2019skynet}.}
\begin{tabular}{| c | c | c | c |c|}
\hline
     Device & Mul. Precision  & Max. GMACs & \# of DSPs & Accuracy \\\hline
     Xilinx      & <9-bit, 11-bit> & 90 &360 & 72.7\% \\
     Ultra96     & <9-bit, 10-bit> & 180  &  & 71.2\% \\
                 & <8-bit, 11-bit> & 180  & & 68.8\% \\
                 & <8-bit, 10-bit> & 180  & & 68.0\% \\\hline
     Intel       & <9-bit, 9-bit> & 180   & 240 & 70.7\%  \\
     5AGXA1      & <12-bit, 12-bit> & 120 & & 72.9\% \\\hline
     
\end{tabular}
\label{tab:exp-dsp-precision}
\vspace{-12pt}
\end{table}

\begin{table}
\centering
\small
\def\arraystretch{0.9}
\caption{Accelerator performance of SkyNet \cite{zhang2019skynet} under different input image size (after resizing) on Xilinx Ultra96.}
\begin{tabular}{| c | c | c | c |}
\hline
     Input Size & FM precision  & Latency & Accuracy \\\hline
     $320\times 160$ & 9-bit & 40ms & 72.7\% \\
     $340\times 180$ & 9-bit & 65ms & 72.8\% \\ \hline
     $320\times 160$ & 8-bit & 33ms & 68.8\% \\
     $340\times 180$ & 8-bit & 49ms & 69.0\% \\
\hline
\end{tabular}
\label{tab:exp-bram-size}
\vspace{-10pt}
\end{table}

\begin{table}[]
    \centering
    \small
    \def\arraystretch{0.9}
    \caption{Different DNNs generated by our co-design framework with different performance constraints and input image resolutions on Xilinx UltraScale+ ZCU102.}
    \begin{tabular}{|c|c|c|c|}
    \hline
      & \multicolumn{3}{c|}{Target Performance} \\\cline{2-4}
      & 15 FPS & 20 FPS & 30 FPS \\\hline
     $400\times 400$   &  Bundle 5    & Bundle 4    & Bundle 4    \\
                        &  13 Replication      & 14 Replication      & 13 Replication     \\
                        &  Max. 1264 ch & Max. 1008 ch & Max. 1024 ch  \\
                        &  mAP 46.1    & mAP 42.4    & mAP 43.9    \\
                        \hline
     $300\times 300$    &  Bundle 1    & Bundle 1    & Bundle 5    \\
                        &  15 Replication    & 14 Replication   & 15 Replication      \\
                        &  Max. 1120 ch & Max. 784 ch & Max. 736 ch  \\
                        &  mAP 45.4    & mAP 44.3    & mAP 39.7    \\
     \hline
     
     \multicolumn{4}{|l|}{
     \footnotesize Bundle 1: conv\_3x3\_stride1}\\
     \multicolumn{4}{|l|}{
     \footnotesize Bundle 2: conv\_5x5\_stride1} \\
     \multicolumn{4}{|l|}{
     \footnotesize Bundle 3: conv\_3x3\_stride1 + conv\_5x5\_stride1}\\
     \multicolumn{4}{|l|}{
     \footnotesize Bundle 4: dw-conv\_3x3\_stride1 + conv\_1x1}\\
     \multicolumn{4}{|l|}{
     \footnotesize Bundle 5: dw-conv5x5\_stride1 + conv1x1
     }\\
     \hline
    \end{tabular}
    \label{tab:dnns-on-zcu102}
    \vspace{-8pt}
\end{table}

We first demonstrate that FPGA-oriented IP and DNN design will have a large impact on accelerator performance.
We use SkyNet~\cite{zhang2019skynet}, a light-weight object detection network, as the baseline.
Table \ref{tab:exp-dsp-precision} shows that different data precisions result in 30\% to 50\% difference in peak performance under 250MHz on Xilinx FPGA and Intel FPGA, while the accuracy does not change dramatically.
It implies that the data precision sometimes is a more sensitive design factor in FPGA accelerator than in DNN model.
Exploiting device-oriented NAIS co-design can take advantage of such difference in sensitivity and come up with DNNs that best match the hardware.

Table \ref{tab:exp-bram-size} shows another example regarding BRAM consideration in Xilinx Ultra96 FPGA using RAMB18E1.
It shows that when the precisions of feature map data are the same, the model accuracy shows negligible difference but the latency shows 32\% and 38\% difference between two input image sizes. This is because when the image is resized to $340 \times 180$, the total bit number of one image tile exceeds 18Kb and occupies two memory blocks, while being resized to $320 \times 160$, one image tile (following the same tiling rule) only consumes one memory block.
Besides the computation capacity, the $340\times 180$ input results in less efficient BRAM utilization and more off-chip data movements, and thus longer latency.

We then apply our NAIS methodology on an object detection task on FPGA for autonomous driving under different input image resolutions and latency constraints.
The target device is Xilinx UltraScale+ ZCU102, a large scaled FPGA with 599,550 logic cells, 32.1Mb block RAM and 2,520 DSP slices.
We set the performance requirements to be 15 FPS, 20 FPS and 30 FPS, respectively, corresponding to different driving speeds in busy downtown, urban street and highway. 
We also consider two input resolutions, $400\times 400$ and $300\times 300$, respectively.
As shown in Table \ref{tab:dnns-on-zcu102},
under each constraint and input resolution, our co-design engine proposes a DNN that is built by replicating a pre-optimized Bundle, as described in Section~\ref{sec:overall-co-design-flow}.
In each scenario, we show the Bundle used for building the DNN, as well as the number of replications and maximum number of channels.
The DNNs are trained and tested on a subset of VOC 2012 dataset, including bike, car, bus and person, which are most related to autonomous driving.
It shows that with different inputs and target performance, the generated DNNs are different. For example, when the input resolution is $400 \times 400$, more light-weight depth-wise Bundles are selected such as Bundle 4 and 5; when the input resolution is $300 \times 300$, Bundle 1 seems more preferable.
This result implies that such a co-design is helpful in searching for the best DNNs within performance constraints under varied circumstances.

\begin{table}
\centering
\small
\def\arraystretch{0.9}
\setlength{\tabcolsep}{4pt}
\caption{Inference latency (ms) of DNNs on different GPUs \cite{pytorch-github}.}
\begin{tabular}{| c | c | c | c | c | c | c |}
\hline
     
    DNN     & \multicolumn{2}{c|}{Titan V}
            & \multicolumn{2}{c|}{1080 Ti}
            & \multicolumn{2}{c|}{2080 Ti} \\\cline{2-7}
     Models & Single & Half & Single & Half & Single & Half \\\hline
    Densenet121 & 17.49 & 11.87 & 23.53 & 18.62 & 16.70 & 13.47 \\
    Densenet161 & 39.33 & 22.88 & 51.53 & 42.26 & 34.64 & 27.10 \\
    Densnet169 & 23.63 & 16.04 & 31.82 & 25.27 & 21.94 & 17.54 \\
    Densnet201 & 30.93 & 20.70 & 47.73 & 33.01 & 28.89 & 22.51 \\
    Resnet18 & 4.82 & 3.09 & 6.43 & 5.65 & 4.89 & 3.42\\ 
    Resnet34 & 8.43 & 5.12 & 10.97 & 9.77 & 8.62 & 5.67 \\
    Resnet50 & 14.27 & 7.61 & 20.17 & 16.26 & 14.65 & 9.04 \\
    Resnet101 & 23.96 & 12.80 & 33.02 & 27.49 & 24.57 & 14.51 \\
    Resnet152 & 34.22 & 18.11 & 47.02 & 38.88 & 35.15 & 20.52 \\
    Vgg16 & 22.94 & 10.96 & 33.73 & 30.69 & 23.70 & 16.14 \\
    Vgg19 & 27.55 & 12.72 & 39.95 & 36.71 & 28.03 & 17.89 \\

\hline
\end{tabular}
\label{tab:gpu-models}
\end{table}

\begin{table}[]
    \centering
    \small
    \def\arraystretch{0.9}
    \setlength{\tabcolsep}{2.0pt}
    \caption{Inference performance on embedded GPU Jetson TX2.}
    \begin{tabular}{|c|c|c|| c | c| }
    \hline
                & $1280 \times 720$ & $320 \times 160$ & Accuracy & Dataset  \\\hline
    YOLO v3     &  3.3 fps & 12.7 fps &  51.5 @ $320\times 320$ ($mAP_{50}$) & COCO \\
    SkyNet      &  20.5 fps  &  67.3 fps & 73.1 @ $320\times 160$ (IoU) & DAC-SDC \cite{DAC-SDC-dataset} \\
    
    \hline
    \end{tabular}
    
    \label{tab:yolo-skynet}
\end{table}

For GPU platform,
we first show a summary of popular DNN models regarding their inference latency on various GPU platforms, including Titan V, 1080 Ti and 2080 Ti.
The summary is shown in Table \ref{tab:gpu-models},
where part of the data are obtained from open-source repository \cite{pytorch-github}.
The input images are $224 \times 224 \times 3$ with a batch size of 16 with single and half precision.
In addition to powerful GPUs, we also make a performance comparison between YOLO v3 \cite{redmon2018yolov3} to SkyNet \cite{zhang2019skynet} on an embedded GPU, Nvidia Jetson TX2,
where SkyNet is showing appealing real-time performance.
SkyNet is a light-weight object detection network we proposed that won the 2019 DAC-SDC competition \cite{DAC-SDC}.
It is composed of basic Bundles, where each Bundle has a depth-wise $3\times 3$ convolution layer followed by a point-wise $1\times 1$ convolution layer.
Instead of a traditional top-down design method which starts from a large DNN and prunes it till reaching required performance, SkyNet was designed by utilizing our proposed NAIS idea discussed in Section \ref{sec:overall-co-design-flow}. 

Our SkyNet design on Jetson TX2 is an initial demonstration of the potential of such NAIS approach.
As a future research direction, GPU implementation shall be optimized during NAIS.

%% file: sections/06-related.tex
\section{related work}


%
For FPGA-based DNN implementations, technologies such as quantization~\cite{qiu2016going, cheng2019uL2Q} and model compression~\cite{han2017ese} are used to reduce DNN model size, while FPGA resource allocation~\cite{Xiaofan2017High} and fine-grained pipeline architecture~\cite{zhang2018dnnbuilder} are proposed to deliver low latency accelerators.
Other works explore FPGA accelerator parameter configuration~\cite{motamedi2016design,zhong2017design,clouddnn} and optimizations such as loop unrolling and pipelining, but they do not explore configurations on the DNN side.
Besides, there are works on DNN and FPGA co-design, which explores both DNN model and accelerator designs.
The work in~\cite{kwon2018co} discussed the DNN and accelerators for embedded vision applications. It first designed a specific DNN accelerator targeting SqueezeNet \cite{iandola2016squeezenet},
and then proposed a tailored DNN model called SqueezeNext according to the hardware utilization of different layers of SqueezeNet.
Another work \cite{jiang2019accuracy} proposed a framework named FNAS, which is a reinforcement learning based NAS by combining the estimated FPGA inference latency into the reward function.
However, none of these works applied simultaneous DNN and FPGA implementation search, as NAIS proposed in our work.

On the other hand, for GPU based DNN search and implementation, NAS has seen a big success in designing high quality DNN models that outperform manually designed ones \cite{elsken2019neural}.
Most early NAS works target purely on improving model accuracy,
while recent works have been conducting performance-aware searches by incorporating estimated hardware performance such as inference latency on GPU or CPU into the NAS engine.
%
One representative work~\cite{cai2018proxylessnas} addressed the high memory consumption issue as well as the high computational cost of differentiable NAS, and solved the problem with gradient-based approach to enable hardware-aware neural architecture search.
Another work~\cite{cheng2018searching} discussed device-aware
neural architecture search by extending the NAS into a multiple-objective problem. Targeting difference devices, their framework came up with a Pareto Frontier regarding DNN accuracy and energy. 
Though these works are closely related to DNN and GPU co-design, they missed the opportunities in device-oriented implementation optimizations and their possible guidance to DNN design, which is an essential goal in NAIS.



%% file: sections/07-conandfuwork.tex
\section{conclusions and future works}
In this paper, we proposed a DNN and implementation co-design methodology, called \textbf{N}eural \textbf{A}rchitecture and \textbf{I}mplementation \textbf{S}earch \textbf{(NAIS)}, to explore the opportunities of boosting the development productivity and efficiency of mapping AI algorithms to targeted platforms. 
The NAIS searches DNN models and the underlying hardware implementations simultaneously in a pre-defined co-design space, with the goal of converging to the best hardware specific solution efficiently.
We first demonstrated how NAIS works for DNN/FPGA co-design, and then discussed the NAIS approach for DNN/GPUs co-design. The NAIS approach can generate various design solutions with different accuracy, latency and computing complexity, which helps to find an optimized implementation for application deployment.
We believe that such a NAIS design methodology can benefit the development productivity and the algorithm/hardware system quality for general DNN algorithms. We also provide a detailed application level case study on how autonomous driving can benefit from such a NAIS approach. 
Our future work includes systematic design space definition and application specific full stack optimizations.


\vspace{-8pt}
\section*{acknowledgement}
This work is supported in part by XMotors.ai, Semiconductor Research Corporation (SRC), the IBM-Illinois Center for Cognitive Computing System Research (C3SR) and Advanced Digital Sciences Center (ADSC) in Singapore. The authors would also like to thank Vibhakar Vemulapati for helpful discussions.